\documentclass[11pt,a4paper]{article}
\usepackage[table]{xcolor}
\usepackage[hyperref]{emnlp-ijcnlp-2019}
\usepackage{times}
\usepackage{latexsym}
\usepackage{graphicx}
\usepackage{enumitem}
\usepackage{url}
\usepackage{booktabs}
\usepackage{subfigure}
\usepackage{footnote}
\usepackage{comment}

\usepackage{amsmath}
\usepackage{amsfonts,bm}

\def\eqref#1{equation~\ref{#1}}

\def\1{\bm{1}}

\def\ry{{\textnormal{y}}}

\def\va{{\bm{a}}}
\def\vb{{\bm{b}}}

\def\vd{{\bm{d}}}
\def\ve{{\bm{e}}}

\def\vg{{\bm{g}}}
\def\vh{{\bm{h}}}

\def\vs{{\bm{s}}}

\def\vx{{\bm{x}}}

\def\m1{{\bm{1}}}

\def\mE{{\bm{E}}}

\def\mH{{\bm{H}}}

\def\mU{{\bm{U}}}
\def\mV{{\bm{V}}}
\def\mW{{\bm{W}}}
\def\mX{{\bm{X}}}

\DeclareMathAlphabet{\mathsfit}{\encodingdefault}{\sfdefault}{m}{sl}
\SetMathAlphabet{\mathsfit}{bold}{\encodingdefault}{\sfdefault}{bx}{n}

\newcommand{\softmax}{\mathrm{softmax}}

\newcommand{\sigmoid}{\text{sigmoid}}

\DeclareMathOperator*{\argmax}{arg\,max}

\usepackage{xspace}

\newcommand{\red}[1]{\textcolor{red}{#1}}

\newcommand{\ie}{{\em i.e.,}\xspace}
\newcommand{\eg}{{\em e.g.,}\xspace}

\newcommand{\Na}{({\em a})~}
\newcommand{\Nb}{({\em b})~}

\usepackage{tikz}
\usepackage{amsmath}
\usepackage[hang]{footmisc}
\usepackage{float}
\usepackage{multirow}

\aclfinalcopy

\newcommand{\head}[1]{%
   \textcolor{white}{\textbf{#1}}}

\title{Hierarchical Pointer Net Parsing}
  
\author{
Linlin Liu$^{12}$\thanks{Linlin Liu is under the Joint PhD Program between Alibaba and Nanyang Technological University.} \thanks{Equal contribution.}, Xiang Lin$^1$\footnotemark[2], Shafiq Joty$^{13}$, Simeng Han$^1$, Lidong Bing$^2$\\
$^1$Nanyang Technological University, Singapore\\
$^2$R\&D Center Singapore, Machine Intelligence Technology, Alibaba DAMO Academy\\
$^3$Salesforce Research Asia, Singapore\\
\{linx0057, srjoty, hans0035\}@ntu.edu.sg \\
\{linlin.liu, l.bing\}@alibaba-inc.com
}

\date{}

\begin{document}
\maketitle

\begin{abstract}

Transition-based top-down parsing with pointer networks has achieved state-of-the-art results in multiple parsing tasks, while having a linear time complexity. However, the decoder of these parsers has a sequential structure, which does not yield the most appropriate inductive bias for deriving tree structures. In this paper, we propose  hierarchical pointer network parsers, and apply them to dependency and sentence-level discourse parsing tasks. Our results on standard benchmark datasets demonstrate the effectiveness of our approach, outperforming existing methods and setting a new state-of-the-art.

\end{abstract}

\section{Introduction} \label{sec:intro}

\begin{figure}[t]
\centering
    \includegraphics[scale=0.10]{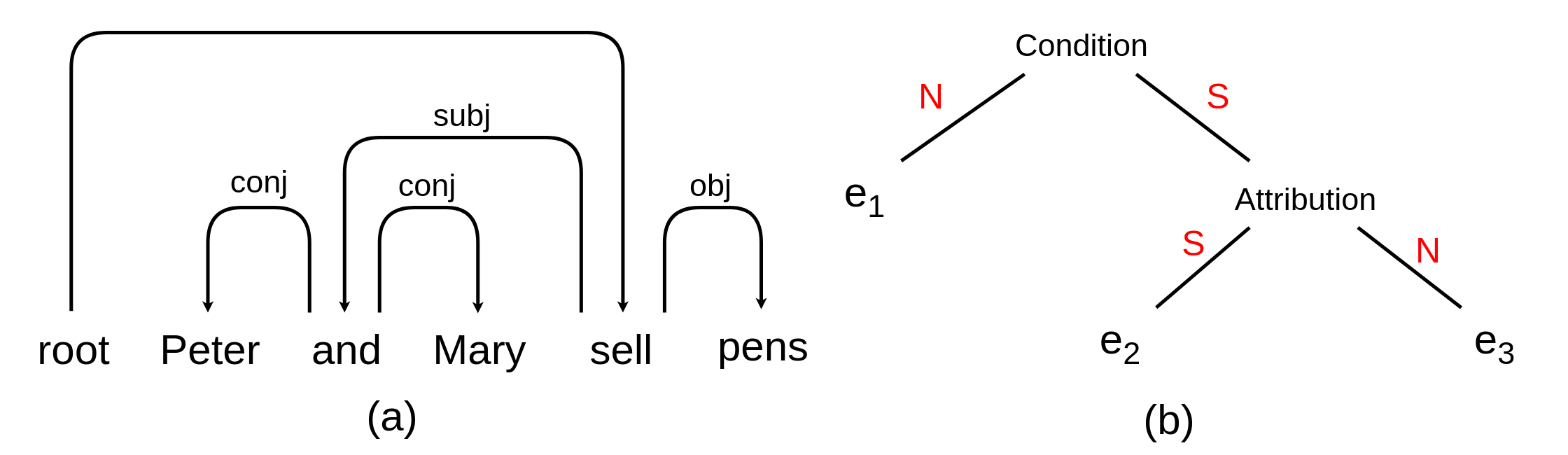} 
\caption{(a) A dependency tree for a sentence; (b) a discourse tree for the sentence ``[Now that's name-dropping,]$_{e_1}$ [if you know]$_{e_2}$ [what I mean.]$_{e_3}$'', where `S' denotes Satellite and `N' denotes Nucleus.}
    \label{fig:example}
\end{figure}

Parsing of sentences is a core natural language understanding task, where the goal is to construct a tree structure that best describes the relationships between the tree  constituents (\eg\ words, phrases). For example, Figure \ref{fig:example} shows examples of a \textbf{dependency tree} and a sentence-level \textbf{discourse tree} that respectively represent how the words and clauses are related in a sentence. Such parse trees are directly useful in numerous NLP  applications, and also serve as intermediate representations for further language processing tasks such as semantic and discourse processing.   

Existing approaches to parsing can be distinguished based on whether they employ a greedy \textbf{transition-based} algorithm  \cite{Marcu99,zhang-nivre-2011-transition,Wang-acl-2017} or a globally optimized algorithm such as \textbf{graph-based} methods for dependency parsing \cite{Eisner:1996:TNP:992628.992688} or \textbf{chart parsing} for discourse \cite{Marcu03,joty-carenini-ng-cl-15}. Transition-based parsers build the tree incrementally by making a series of shift-reduce  decisions. The advantage of this method is that the parsing time is linear with respect to the sequence length. The limitation, however, is that the decisions made at each step are based on local information, disallowing the model to capture long distance dependencies and also causing error propagation to subsequent steps. Recent methods attempt to address this issue using neural networks capable of remembering long range relationships such as Stacked LSTMs \cite{dyer-etal-2015-transition, ballesteros-etal-2015-improved} or using globally normalized models \cite{andor-etal-2016-globally}.


The globally optimized methods, on the other hand, learn scoring functions for subtrees and perform search over all  possible trees to find the most probable tree for a text. Recent graph-based methods use neural models as scoring functions \cite{kiperwasser-goldberg-2016-simple,DozatMann17}. Despite being more accurate than greedy parsers, these methods are generally slow having a polynomial time complexity ($O(n^3)$ or higher).

Recently, transition-based \textbf{top-down} parsing with Pointer Networks \cite{Vinyals_NIPS2015} has attained state-of-the-art results in both dependency and discourse parsing tasks with the same computational efficiency \cite{Xuezhe18,Xiang19}; thanks to the \textbf{encoder-decoder} architecture that makes it possible to capture information from the whole text and the previously derived subtrees, while limiting the number of parsing steps to linear. However, the decoder of these parsers has a {sequential structure}, which may not yield the most appropriate \textbf{inductive bias} for deriving a hierarchical structure. For example, when decoding \emph{``pens''} in Figure \ref{fig:example} in a top-down depth-first manner, the decoder state is directly conditioned on \emph{``and''} as opposed to the states representing its parent \emph{``sell''}. This on one hand may induce irrelevant information in the current state, on the other, as the text length gets longer, the decoder state at later steps tends to forget more relevant information due to long distance. Having an explicit hierarchical inductive bias should allow the model to receive more relevant information and help with the long-term dependency problem by providing shortcuts for gradient back-propagation.

In this paper, we propose a \textbf{Hierarchical Pointer Network (H-PtrNet)} parser to address the above mentioned limitations. In addition to the sequential dependencies, our parser also directly models the parent-child and sibling relationships in the decoding process. We apply our proposed method to both dependency and discourse parsing tasks. To verify the effectiveness of our approach, we conduct extensive experiments and analysis on both tasks. 
Our results demonstrate that in dependency parsing, our model outperforms in most of the languages.
In discourse parsing, we push forward the state-of-the-art in all evaluation metrics. Furthermore, our results on the hardest task of relation labeling have touched human agreement scores on this task. We have released our code at \href{https://ntunlpsg.github.io/project/parser/ptrnet-depparser/}{https://ntunlpsg.github.io/project/parser/ptrnet-depparser/} for research purposes.




\section{Background} \label{sec:background}
\subsection{Dependency Parsing} \label{ssec:background:depanddis}


Dependency parsing is the task of predicting the existence and type of linguistic dependency relations between  words in a sentence (Figure \ref{fig:example}a). Given an input sentence, the output is a tree that shows relationships (\eg\ {\sc{Nominal Subject (NSUBJ)}}, {\sc{Determiner (DET)}}) between \emph{head} words and words that modify those heads, called \emph{dependents} or \emph{modifiers}.

Approaches to dependency parsing can be divided into two main categories:  greedy transition-based parsing and graph-based parsing. In both paradigms, neural models have proven to be more effective than feature-based models where selecting the composition of features is a major challenge. \citet{kiperwasser-goldberg-2016-simple} proposed graph-based and transition-based dependency parsers with a Bi-LSTM feature representation. Since then much work has been done to improve these two parsers. \citet{DozatMann17} proposed a \textbf{bi-affine} classifier for the prediction of arcs and labels based on graph-based model, and achieved state-of-the-art performance. \citet{DBLP:jPTDB} adopted a joint modeling approach by adding a Bi-LSTM POS tagger to generate POS tags for the graph-based dependency parser. Though transition-based methods are superior in terms of time complexity, they fail in capturing the global dependency information when making decisions. To address this issue, \citet{andor-etal-2016-globally} proposed a globally optimized transition-based model. Recently, by incorporating a stack within a pointer network, \citet{Xuezhe18} proposed a transition-based model and achieved state-of-the-art performance across many languages .


\subsection{Discourse Parsing} \label{ssec:background:dis}


{Rhetorical Structure Theory} or {RST} \cite{Mann88} is one of the most influential theories of discourse, which posits a tree structure (called discourse tree) to represent a text (Fig. \ref{fig:example}b). The leaves of a discourse tree represent contiguous text spans called Elementary Discourse Units (EDUs). The adjacent EDUs and larger units are  recursively connected by coherence relations (\eg\ {\sc{Condition}}, {\sc{Attribution}}). Furthermore, the discourse units connected by a relation are distinguished based on their relative importance --- {\sc{Nucleus}} refers to the core part(s) while {\sc{Satellite}} refers to the peripheral one. Coherence analysis in RST consists of two subtasks: \Na identifying the EDUs in a text, referred to as \textbf{Discourse Segmentation}, and \Nb building a discourse tree by linking the EDUs hierarchically, referred to as \textbf{Discourse Parsing}. This work focuses on the more challenging task of discourse parsing assuming that EDUs have already been identified. In fact, state-of-the-art segmenter \cite{Xiang19} has already achieved 95.6 $F_1$ on RST discourse treebank, where the human agreement is 98.3 $F_1$.




Earlier methods have mostly utilized hand-crafted lexical and syntactic features \cite{Marcu03,Feng-14-ACL,joty-carenini-ng-cl-15,Wang-acl-2017}. Recent approaches have shown competitive results with neural models that are able to automatically learn the feature representations in an end-to-end fashion \cite{ji-eisenstein:2014:P14-1,Li-2014-acl}. Very recently, \citet{Xiang19} propose a parser based on pointer networks and achieve state-of-the-art performance. 

\paragraph{\textit{Remark.}} Although related, the dependency and RST tree structures (hence the parsing tasks) are different. RST structure is similar to constituency structure. Therefore, the differences between constituency and dependency structures also hold here.\footnote{There are also studies  that use dependency structure to directly represent the
relations between the EDUs; see \cite{muller-etal-2012-constrained,Li-2014-acl,morey-etal-2018-dependency}.} First, while dependency relations can only be between words, discourse relations can be between elementary units, between larger units or both. Second, in dependency parsing, any two words can be linked, whereas RST allows connections only between two adjacent units. Third, in dependency parsing, a head word can have multiple modifier words, whereas in discourse parsing, a discourse unit can be associated with only one connection. The parsing algorithm needs to be adapted to account for these differences.







\subsection{Pointer Networks.}

Pointer networks \cite{Vinyals_NIPS2015} are a class of encoder-decoder models that can tackle  problems where the output vocabulary depends on the input sequence. They use attentions as pointers to the input elements. An encoder network first converts the input sequence $\mX = (\vx_1, \ldots, \vx_n)$ into a sequence of hidden states $\mH = (\vh_1, \ldots, \vh_n)$. At each time step $t$, the decoder takes the input from previous step, generates a decoder state $\vd_t$, and uses it to attend over the input elements. The attention gives a $\softmax$ distribution over the input elements.


\begin{equation}
s_{t,i} = \sigma(\vd_t, \vh_i); \hspace{1em}
\va_t = \softmax(\vs_t)
\label{eq:attn}
\end{equation}

where $\sigma(.,.)$ is a scoring function for attention, which can be a neural network or an explicit formula like dot product. The model uses $\va_t$ to infer the output: $\hat{\ry}_t = \argmax (\va_t) = \argmax p(\ry_t | \ry_{<t}, \mX, \theta)$
where $\theta$ is the set of parameters. To condition on $\ry_{t-1}$, the corresponding input $\vx_{\ry_{t-1}}$ is copied as the input to the decoder.









\section{Hierarchical Pointer Networks} \label{sec:model}
Before presenting our proposed hierarchical pointer networks, we first revisit how pointer networks have been used for parsing tasks.

\subsection{Pointer Networks for Parsing.}
\citet{Xuezhe18} and \citet{Xiang19} both use a pointer network as the backbone of their parsing models and achieve state-of-the-art performance in dependency and discourse parsing tasks, respectively. As shown in Figures \ref{fig:entire_model} and \ref{fig:entire_model_dis}, in both cases, the parsing algorithm is implemented in a top-down depth-first order. They share the same encoder-decoder structure. A bi-directional Recurrent Neural Network (RNN) encodes a sequence of word embeddings $\mX$ $=$ $(\vx_1, \ldots, \vx_n)$ into a sequence of hidden states $\mH$ $=$ $(\vh_1, \ldots, \vh_n)$. The decoder implements a uni-directional RNN to greedily generate the corresponding tree. It maintains a stack to keep track of the units that yet need to be parsed, \ie\ head words for dependency parsing and larger units for discourse parsing. At each step $t$, the decoder takes out an element from the stack and generates a decoder state $\vd_t$, which is in turn used in the pointer layer to compute the attention over the relevant input elements. In the case of dependency parsing, the representation of the head word is used to find its dependent. For discourse parsing, it uses the representation of the span to identify the break position that splits the text span into two subspans.

{In addition to the tree structure, the parser also deals with the corresponding labelling tasks. Whenever the pointer network yields a newly created pair (\ie head-dependent in dependency parsing, two sub-spans in discourse parsing), a separate classifier is applied to predict the corresponding relation between them.}

\subsection{Limitations of Existing Methods} \label{ssec:limitations}

\begin{figure*}[t]
\centering
\subfigure[Model Structure \label{fig:entire_model}]{\includegraphics[scale=0.13]{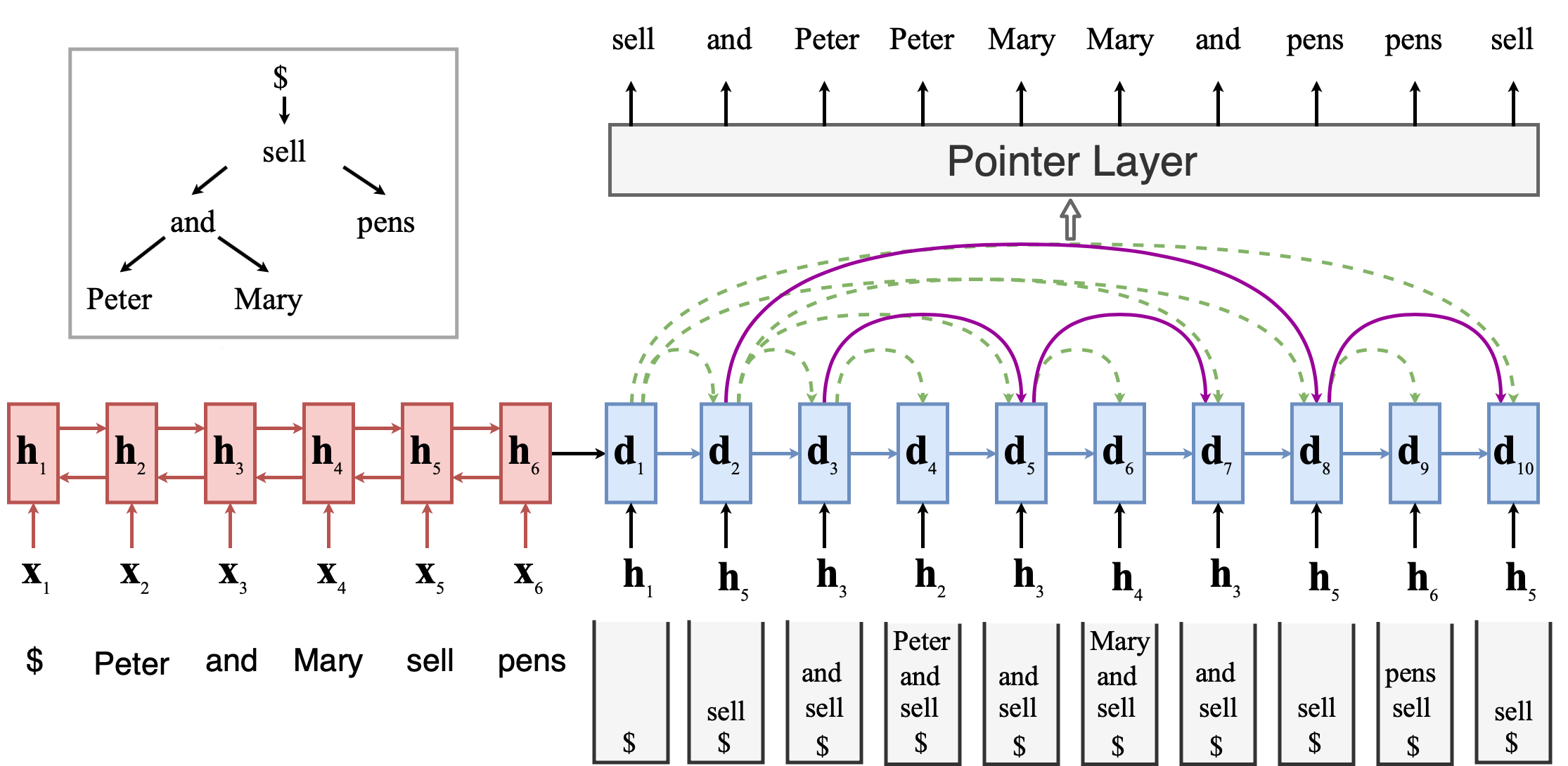}}
\qquad
\subfigure[Decoder Connections \label{fig:dep_decoder_connections}]{\includegraphics[scale=0.5]{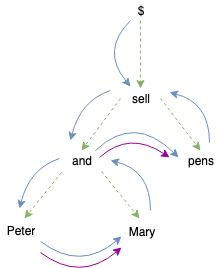}}
\caption{(a) H-PtrNet for dependency parsing. To reduce visual clutter, we do not show the attention scores over all the input elements at each pointing step, rather show only the pointed element. Figure (b) shows the decoder connections in our H-PtrNet model. The StackPointer network of \citet{Xuezhe18} has a sequential decoder (shown as blue straight lines). The Green dash lines indicate parent connections and purple solid lines denote the immediate sibling connections in our model.}
\label{fig:entire_model_dep}
\end{figure*}

One crucial limitation of the existing models is that the decoder has a linear structure, although the task is to construct a hierarchical structure. This can be noticed in the Figures \ref{fig:entire_model_dep} and \ref{fig:entire_model_dis}, where the current decoder state $\vd_t$ is conditioned on the previous state $\vd_{t-1}$ (see horizontal blue lines), but not on its parent's decoder state or siblings' decoder state, when it was pointed from its head. This can induce irrelevant information if the previous decoding state corresponds to an element that is not relevant to the current element. For example, in Figure \ref{fig:entire_model_dep}, the decoder state for pointing to ``pens'' is conditioned on the state used for pointing to ``and'', but not the one used for pointing to ``sell'', which is more relevant according to the dependency structure. Also, the decoder state for ``sell'' is far apart from the one for ``pens''. Therefore, more relevant information could be diminished in a sequential decoder, especially for long range dependencies. 





\subsection{Hierarchical Decoder} \label{ssec:difference}

To address the above issues, we propose hierarchical pointer network (H-PtrNet), which poses a hierarchical decoder that reflects the underlying tree structure. H-PtrNet has the same encoder-decoder architecture as the original pointer network except that  each decoding state $\vd_t$ is conditioned directly on its parent's decoder state $\vd_{p(t)}$ and its immediate sibling's decoder state $\vd_{s(t)}$ in addition to the previous decoder state $\vd_{t-1}$ and parent's encoder state $\vh_{p(t)}$ (from input). Formally, the pointing mechanism in H-PtrNet can be defined as:

\begin{eqnarray}
\vd_{t} = f(\vd_{p(t)}, \vd_{s(t)}, \vd_{t-1}, \vh_{p(t)})
\label{eq:hptr1} \\
s_{t,i} = \sigma(\vd_t, \vh_i) \label{eq:sc}
\end{eqnarray}
\begin{eqnarray}
p(\ry_t | \ry_{<t}, \mX, \theta) \hspace{-0.3em} = \softmax(\vs_t) \hspace{-0.3em} = \frac{\exp(s_{t,i})}{\sum_i \exp(s_{t,i})}
\label{eq:hptr}  
\end{eqnarray}

\noindent where $f(.)$ is a fusion function to combine the four components into a decoder state, and other terms are similarly defined as before for Eq. \ref{eq:attn}. Figure \ref{fig:dep_decoder_connections} shows an example of H-PtrNet  decoder  connections  for dependency parsing.

The fusion function $f(.)$ can be implemented in multiple ways and may depend on the specific parsing task. More variants of the fusion function will be discussed in Section \ref{sec:experiments}. 



\paragraph{\textbf{Decoder Time Complexity.}} Given a sentence of length $n$, the number of decoding steps to build a parse tree is linear. The attention mechanism at each decoding step computes an attention vector of length $n$. The overall decoding complexity is $O(n^2)$, which is same as the StackPointer Parser \cite{Xuezhe18}.

\paragraph{\textit{Remark.}} If we look at the decoding steps of the StackPointer Parser \cite{Xuezhe18} more closely, we notice that it also takes the decoder state of the immediate sibling (when it points to itself). This decoder state represents the state after all its children are generated. Thus it contains information about its children. In contrast, in our model we consider the decoder state when the sibling was first generated from its parent. Therefore this state contains the sibling's parent information, which helps with capturing long term dependencies.  


\subsection{Model Specifics for Dependency Parsing} \label{ssec:dependency}

Figure \ref{fig:entire_model} shows the encoding and decoding steps of H-PtrNet for dependency parsing. We use the same encoder as \citet{Xuezhe18} (red color).\footnote{https://github.com/XuezheMax/NeuroNLP2} Given a sentence, a convolutional neural network (CNN) is used to encode character-level representation of each word, which is then concatenated with word embedding and POS embedding vectors to generate the input sequence $\mX = (\vx_1, \ldots, \vx_n)$. Then a three-layer bi-directional LSTM encodes $\mX$ into a sequence of hidden states  $\mH = (\vh_1, \ldots, \vh_n)$. The decoder (blue color) is a single layer uni-directional LSTM, and also maintains a stack to track of the decoding status. At each decoding step $t$, the decoder receives the \emph{encoder state} of the parent from the stack. In addition, it gets \emph{decoder states} from three different sources: previous decoding step $\vd_{t-1}$, parent $\vd_{p(t)}$ and immediate sibling $\vd_{s(t)}$.

{Instead of simply feeding these three components to the decoder, we incorporate a gating mechanism to generalize the ability of our model to extract the most useful information. Eventually,}
the \textbf{fusion function} $f(\vd_{p(t)}, \vd_{s(t)}, \vd_{t-1}, \vh_{p(t)})$ in Eq. \ref{eq:hptr1} is defined with a \emph{gating mechanism}. We experimented with two different gating functions:


\begin{multline}
\vg_{t} = \sigmoid(\mW_{gd}\vd_{t-1}+\mW_{gp}\vd_{p(t)} \\ +\mW_{gs}\vd_{s(t)} + \vb_g)
\label{eq:hptr_gate1}
\end{multline}
\begin{multline}
\vg_{t} = \sigmoid(\mW_{gp}(\vd_{t-1} \odot \vd_{p(t)}) \\ +\mW_{gs}(\vd_{t-1} \odot \vd_{s(t)}) + \vb_g)
\label{eq:hptr_gate2}
\end{multline}

\noindent where $\mW_{gp}$, $\mW_{gs}$, $\mW_{gd}$ and $\vb_g$ are the gating weights. The fusion function is then defined as

\begin{eqnarray}
\vh_{t}' \hspace{-0.7em} &=& \hspace{-0.7em} \tanh ( \mW_{d}\vd_{t-1} + \mW_{p}\vd_{p(t)} + \mW_{s}\vd_{s(t)})
\label{eq:hptr_hiddenstate1} \\
\vh_{t}'' \hspace{-0.7em} &=& \hspace{-0.7em} \vg_{t} \odot \vh_{t}'
\label{eq:hptr_hiddenstate2}\\
\vd_{t} \hspace{-0.7em} &=& \hspace{-0.7em} \text{LSTM}(\vh_{t}'', \vh_{p(t)})
\label{eq:hptr_hiddenstate3}
\end{eqnarray}

\noindent where $\mW_{d}$, $\mW_{p}$, $\mW_{s}$ are the weights to get the intermediate hidden state $\vh_{t}'$, and $\vg_t$ is a gate to control the information flow from the three decoder states 
$\text{LSTM}$ is the LSTM layer that accepts $\vh_{t}''$ as the the hidden state and $\vh_{p(t)}$ as its input. The LSTM decoder state $\vd_{t}$ is then used to compute the attention distribution over the  encoder  states in pointer layer.

\paragraph{Pointer and Classifier.} 

Same as \citet{Xuezhe18}, the pointer and the label classifier are implemented as \textbf{bi-affine} layers. Formally, the scoring function $\sigma(\vd_t, \vh_i)$ in Eq. \ref{eq:sc} is defined as:

\begin{multline}
s_{t,i} = g_1(\vd_t)^T\mW g_2(\vh_i) + \mU g_1(\vd_t) +\\ \mV g_2(\vh_i) + \vb
\label{eq:biaffine_score}
\end{multline}

\noindent where $\mW$, $\mU$ and $\mV$ are the weights, and $g_1(.)$ and $g_2(.)$ are two single layer MLPs with ELU activations. The dependency label classifier has the same structure as the pointer. More specifically, 

\begin{multline}
p(\ry_l|\mX) = \softmax (g'_1(\vd_t)^T \mW_c g'_2(\vh_{k}) +\\ \mU_cg'_1(\vd_t) + \mV_c g'_2(\vh_{k}) + \vb_c)
\label{eq:biaffine_classifier}
\end{multline}

\noindent where $\vh_k$ is the encoder state of the dependent word,  $\vd_t$ is the decoder state of the head word, $\mW_c$, $\mU_c$ and $\mV_c$ are the  weights, and $g'_1(.)$ and $g'_2(.)$ are two single layer MLPs with ELU activations. 


\paragraph{Partial Tree Information.} Similar to \citet{Xuezhe18}, we provide the decoder at each step with higher order information about the parent and the sibling of the current node.

\subsection{Model Specifics for Discourse Parsing.} \label{ssec:discourseparsing}

\begin{figure}[t!]
\centering
\includegraphics[scale=0.11]{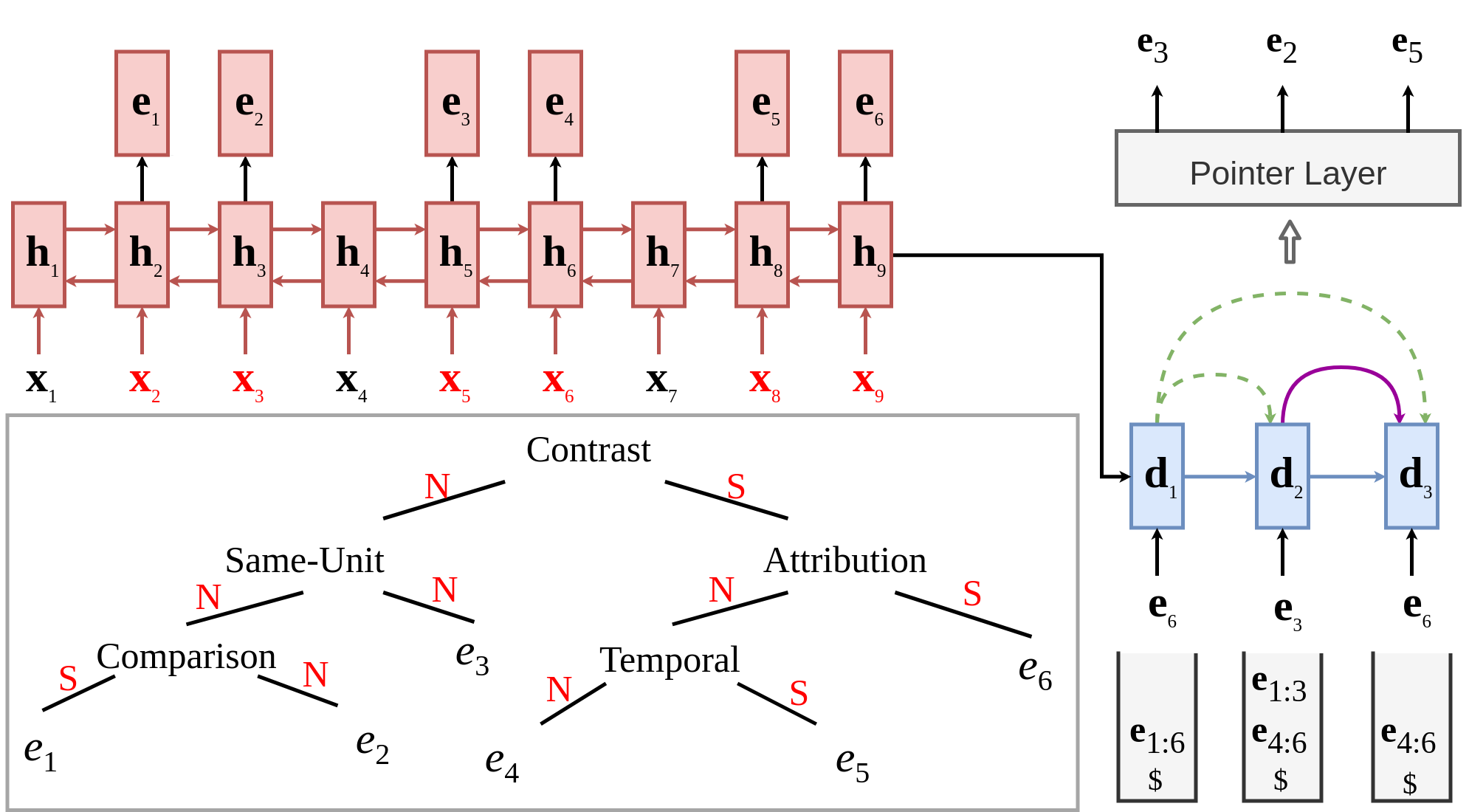}
\caption{H-PtrNet for discourse parsing. The input symbols in red ($\mathbf{x}_2, \mathbf{x}_3, \mathbf{x}_5, \mathbf{x}_6, \mathbf{x}_8, \mathbf{x}_9$) represent last words of the respective EDUs. To avoid visual clutter, we do not show the attention distributions over the EDUs, rather only show the decisions. Green dash lines indicate parent connections and purple solid lines denote the immediate sibling connections.}
\label{fig:entire_model_dis}
\end{figure}

For discourse parsing, our model uses the same structure as \citet{Xiang19}.\footnote{\url{https://ntunlpsg.github.io/project/parser/pointer-net-parser}} The encoder is a 5-layer bidirectional RNN based on Gated Recurrent Units (BiGRU) \cite{ChoGRU}. As shown in Figure \ref{fig:entire_model_dis}, after obtaining a sequence of encoder hidden states representing the words, the  last hidden states of the EDUs (\eg $\vh_2$, $\vh_3$, $\vh_5$, $\vh_6$, $\vh_8$ and $\vh_{10}$) are taken as the EDU representations, generating a sequence of EDU representations $\mE = (\ve_1, \ldots, \ve_m)$ for the input sentence. 

Our hierarchical decoder is based on a 5-layer unidirectional GRU. The decoder maintains a \textbf{stack} to keep track of the spans that need to be parsed further. At time step $t$, the decoder takes the text span (\eg $\ve_{i:j}$) representation from the top of the stack and receives the corresponding {parent decoder state} $\vd_{p(t)}$, {sibling decoder state} $\vd_{s(t)}$ and {previous decoder state} $\vd_{t-1}$ as the input to generate a current decoder state $\vd_t$. For discourse parsing, we apply Eq. \ref{eq:hptr_hiddenstate1} and \ref{eq:hptr_hiddenstate3} (with GRU) to implement the fusion function $f(.)$ and to get the decoder state $\vd_t$.\footnote{Adding gating mechanism did not give any gain, rather increased the number of parameters.} The decoder state is then used in the pointer layer to compute the attention score over the current text span (\eg $\ve_{i:j}$) in order to find the position $k$ to generate a new split $(\ve_{i:k},\ve_{k+1:j})$. The parser then applies a relation classifier $\Phi(\ve_{i:k},\ve_{k+1:j})$ to predict the relation and the nuclearity labels for the the new split.

For pointing, the parser uses a simple \textbf{dot product} attention. For labeling, it uses a \textbf{bi-affine classifier} similar to the one in Eq. \ref{eq:biaffine_classifier}. It takes the representations of two spans (\ie $\ve_k$ for $\ve_{i:k}$, $\ve_j$ for $\ve_{k+1:j}$) as input and predicts the corresponding relation between them. Whenever the length of any of the newly created span ($\ve_{i:k}$ and $\ve_{k+1:j}$) is larger than two, the parser pushes it onto the stack for further processing. Similar to dependency parsing, the decoder is also provided with \textbf{partial tree information} -- the representations of the \textbf{parent} and the immediate \textbf{sibling}.

\subsection{Objective Function}
Same as \citet{Xuezhe18} and \citet{Xiang19}, our parsers are trained to minimize the total $\log$ loss (cross entropy) for building the right tree \textbf{structure} for a given sentence $\mX$. The structure loss $\mathcal{L}$ is the pointing loss for the pointer network:

\begin{eqnarray}
\mathcal{L}(\theta) = 
\displaystyle - \sum_{t=1}^{T} \log P_{\theta}(\ry_t | \ry_{<t}, \mX)  \label{eq:LossParser}
\end{eqnarray}

\noindent where $\theta$ denotes the model parameters, $y_{<t}$ represents the subtrees that have been generated by our parser at previous steps, and $T$ is the number needed for parsing the whole sentence (\ie number of words in dependency parsing and spans containing more than two EDUs in discourse parsing). The label classifiers are trained simultaneously, so the final loss function is the sum of structure loss (Eq. \ref{eq:LossParser}) and the loss for label classifier.

\section{Experiments} \label{sec:experiments}
In this section, we describe the experimental details about dependency parsing and discourse parsing, as well as the analysis on both tasks.

Apart from the two gating-based fusion functions described in Section \ref{sec:model} (Eq. \ref{eq:hptr_gate1}-\ref{eq:hptr_gate2}), we experimented with three different versions of our model depending on which connections are considered in the decoder. We append suffixes \textbf{P} for \emph{parent}, \textbf{S} for  \emph{sibling} and \textbf{T} for \emph{temporal} to the model name (H-PtrNet) to denote different versions.


\begin{itemize}[leftmargin=*]
    \item \textbf{H-PtrNet-P}: The H-PtrNet model with fusion function $\vd_{t} = f(\vd_{p(t)}, \vh_{p(t)})$, where the decoder receives hidden (decoder) state only from the parent ($\vd_{p(t)}$) in each decoding step. Note that $\vh_{p(t)}$ is the encoder state of the parent.
    
    \item \textbf{H-PtrNet-PS}: The H-PtrNet model with fusion function $\vd_{t} = f(\vd_{p(t)}, \vd_{s(t)}, \vh_{p(t)})$, where the decoder receives the hidden states from both the parent and sibling in each decoding step.
    
    \item \textbf{H-PtrNet-PST}: This is the full model with fusion function $\vd_{t} = f(\vd_{p(t)}, \vd_{s(t)}, \vd_{t-1}, \vh_{p(t)})$ (Eq. \ref{eq:hptr1}). In this model, the  decoder receives the hidden states from its parent, sibling and previous step in each decoding step.

\end{itemize}


\subsection{Dependency Parsing}
\paragraph{Dataset.}
We evaluate our model on the English Penn Treebank (PTB v3.0) \cite{Marcus93}, which is converted to Stanford Dependencies format with Stanford Dependency Converter 3.3.0 \cite{Schuster2016EnhancedEU}.
To make a thorough empirical comparison with previous studies, we also evaluate our system on seven (7) languages from the Universal Dependency (UD) Treebanks\footnote{http://universaldependencies.org/} (version 2.3).

\paragraph{Metrics.}
We evaluate the performance of our models with unlabeled attachment score (UAS) and labeled attachment score (LAS). We ignore punctuations in the evaluation for English.

\renewcommand{\arraystretch}{1.3}
\begin{savenotes}
\begin{table*}[t!]
\centering
\scalebox{0.8}{\begin{tabular}{c|cc|cc|cc|cc}
\setlength{\arrayrulewidth}{1mm}
\setlength{\tabcolsep}{20pt}
\renewcommand{\arraystretch}{1.5}
& \multicolumn{2}{c}{\textbf{StackPtr (code)}} & \multicolumn{2}{c}{\textbf{H-PtrNet-PST (Gate)}} & \multicolumn{2}{c}{\textbf{H-PtrNet-PST (SGate)}} \\
\midrule
& UAS & LAS & UAS & LAS & UAS & LAS \\
\midrule
\bf{bg} & 94.17$\pm$0.11 & 90.63$\pm$0.06 & 94.20$\pm$0.16 & 90.70$\pm$0.14 & \bf{94.50$\pm$0.16} & \bf{91.01$\pm$0.20} \\
\midrule
\bf{ca} & \bf{93.82$\pm$0.06} & \bf{91.99$\pm$0.07} & 93.78$\pm$0.03 & 91.92$\pm$0.03 & 93.67$\pm$0.06 & 91.82$\pm$0.07 \\
\midrule
\bf{en} & 90.97$\pm$0.07 & 89.06$\pm$0.08 & \bf{91.03$\pm$0.19} & \bf{89.07$\pm$0.16} & 90.94$\pm$0.12 & 89.01$\pm$0.12 \\

\midrule
\bf{de} & 87.97$\pm$0.20 & 83.75$\pm$0.21 & \bf{88.14$\pm$0.22} & \bf{83.89$\pm$0.26} & 88.06$\pm$0.17 & 83.83$\pm$0.13 \\
\midrule
\bf{fr} & 91.57$\pm$0.23 & 88.76$\pm$0.21& 91.63$\pm$0.15 & 88.70$\pm$0.14 & \bf{91.69$\pm$0.07} & \bf{88.80$\pm$0.11}  \\
\midrule
\bf{it} & 93.76$\pm$0.13 & 92.00$\pm$0.08 & 93.73$\pm$0.08 & 91.90$\pm$0.10 & \bf{93.88$\pm$0.05} & \bf{92.09$\pm$0.02} \\
\midrule
\bf{ro} & 91.15$\pm$0.12 & 85.54$\pm$0.13 & \bf{91.34$\pm$0.18} & \bf{85.73$\pm$0.22} & 91.09$\pm$0.09 & 85.36$\pm$0.10 \\
\bottomrule
\end{tabular}
}
\caption{Dependency parsing results on 7 UD Treebanks. \textbf{StackPtr (code)} denotes the experiments we rerun on our machine. \textbf{H-PtrNet-PST (Gate)} and \textbf{H-PtrNet-PST (SGate)} are H-PtrNet models with gating mechanism.}
\label{table:dep_result}
\end{table*}
\end{savenotes}

\paragraph{Experimental Settings.}
We use the same setup as \citet{Xuezhe18} in the experiments for English Penn Treebank and UD Treebanks. For a fair comparison, we rerun their model with the hyperparameters provided by the authors on the same machine as our experiments. For all the languages, we follow the standard split for training, validation and testing. It should be noted that \citet{Xuezhe18}  used UD Treebanks 2.1, which is not the most up-to-date version. Therefore, during experiments, we rerun their codes with UD Treebanks 2.3 to match our experiments. To be specific, we use  structured-skipgram \cite{ling2015two} for English and German, while Polyglot embedding  \cite{al2013polyglot} for the other languages.  Adam optimizer \cite{Kingma2015AdamAM} is used as the optimization algorithm.  We apply 0.33 dropout rate between layers of encoder and to word embeddings as well as Eq. \ref{eq:hptr_hiddenstate2}. We use beam size of 10 for English Penn Treebank, and beam size of 1 for UD Treebanks.  The gold-standard POS tags is used for English Penn Treebank. We also use the universal POS tags \cite{petrov12} provided in the dataset for UD Treebanks. See Appendix for a complete list of hyperparameters.

\paragraph{Results on UD Treebanks.}

We evaluate on 7 different languages from the UD Treebanks: 4 major ones:  English (en), German (de), French (fr), and Italian (it), and 3 relatively minor ones: Bulgarian (bg), Catalan (ca), and Romanian (ro). Table \ref{table:dep_result} shows the results. We refer to the results of our run of the code released by \citet{Xuezhe18} as StackPtr (code).\footnote{We do not directly report the results from their paper because we use a different version of the UD Treebanks.} 
StackPtr (code) and our models are trained in identical settings making them comparable. H-PtrNet-PST (Gate) (Eq. \ref{eq:hptr_gate1}) and H-PtrNet-PST (SGate) (Eq. \ref{eq:hptr_gate2}) are H-PtrNet models with gating mechanism. Element wise product in Eq. \ref{eq:hptr_gate2} has the effect of similarity comparison, so we denote it as {\sc \textbf{SGate}}. With gating mechanism, our model shows consistent improvements against the baseline on bg, en, de, fr, it and ro. We also tested H-PtrNet-PS on these 7 languages, but the performances are worse than StackPtr.



\renewcommand{\arraystretch}{1.2}
\begin{table}[t!]
\centering
\scalebox{0.80}{\begin{tabular}{l|ccc}  
\textbf{Approach} & \multicolumn{1}{c}{\bf{UAS}} & \bf{LAS}\\
\midrule
\rowcolor{black!60}
\head{\bf{Baselines}} & & \\
StackPtr (paper) & 96.12$\pm$0.03& 95.06$\pm$0.05 \\ 
StackPtr (code)  & 95.94$\pm$0.03 & 94.91$\pm$0.05 \\
%
\midrule
\rowcolor{black!60}
\head{\bf{Proposed Model}} & & \\
H-PtrNet-PST (Gate) & 96.03$\pm$0.02 & 94.99$\pm$0.02 \\
H-PtrNet-PST (SGate) & 96.04$\pm$0.05 & 95.00$\pm$0.06 \\
H-PtrNet-PS (Gate) & \bf{96.09$\pm$0.05} & \bf{95.03$\pm$0.03} \\
\bottomrule
\end{tabular}}
\caption{Dependency parsing results on English Penn Treebank v3.0.}
\label{table:dep-english}
\end{table}

\paragraph{Results on English Penn Treebank.}
Table \ref{table:dep-english} presents the results on English Penn Treebank. StackPtr (paper) refer to the results reported by \citet{Xuezhe18}, and StackPtr (code) is our run of their code in identical settings as ours. Our model H-PtrNet-PST (Gate) outperforms the baseline by 0.09 and 0.08 in terms of UAS and LAS, respectively. Performance of H-PtrNet-PST (SGate) is close to that of H-PtrNet-PST (Gate), though we see slight improvement. We also test H-PtrNet-PS (Gate), the model with parent and sibling connections only, which further improves the performance to 96.09 and 95.03 in UAS and LAS.


\paragraph{Performance Analysis.}

\begin{figure}[t]
\hspace{-1em}
\subfigure[H-PtrNet-PST (no gating) \label{fig:dependency_performance_1}]{\includegraphics[scale=0.252]{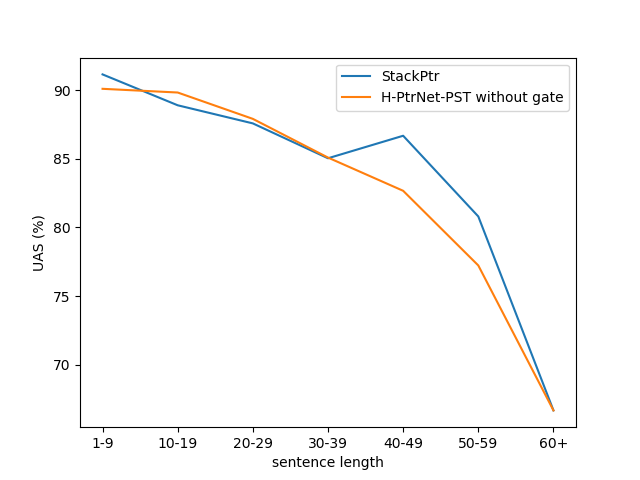}}
\hspace{-1em}
\subfigure[H-PtrNet-PST (SGate)\label{fig:dependency_performance_2}]{\includegraphics[scale=0.252]{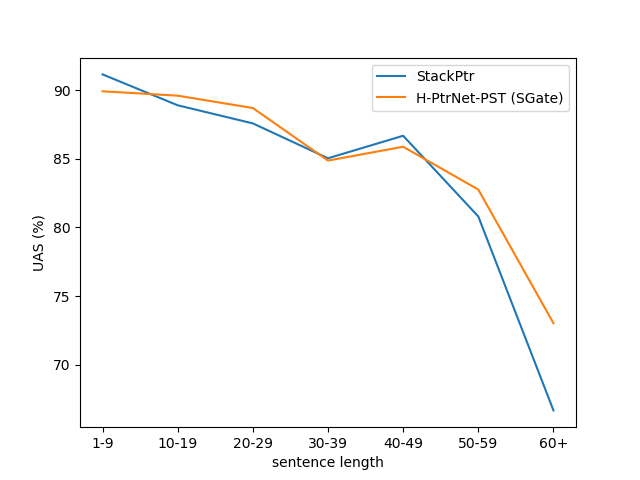}}
\caption{UD German parsing performance in terms of sentence length.}
\label{fig:dependency_performance}
\end{figure}

To make a thorough analysis of our model, we breakdown UAS in terms of sentence lengths to compare the performance of our model and StackPtr. We first take the performance on UD German as an example, which is shown in Figure \ref{fig:dependency_performance}. The blue line shows the performance of StackPtr, and the orange line shows the performance of our model. From Figure \ref{fig:dependency_performance_1} we can see that our model without gate performs better on relatively short sentences (10 to 29 words), however, the accuracy drops on longer sentences. {The reason could be that adding parent and sibling hidden states to decoder may amplify error accumulation from early parsing mistakes.}

\vspace{1em}

\begin{figure}[t]
\hspace{-1em}
\subfigure[UD French \label{fig:dep_performance_fr}]{\includegraphics[scale=0.252]{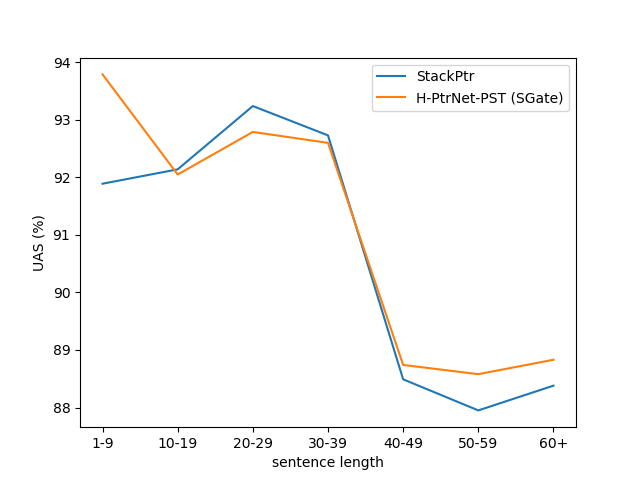}}
\hspace{-1em}
\subfigure[UD Italian
\label{fig:dep_performance_it}]{\includegraphics[scale=0.252]{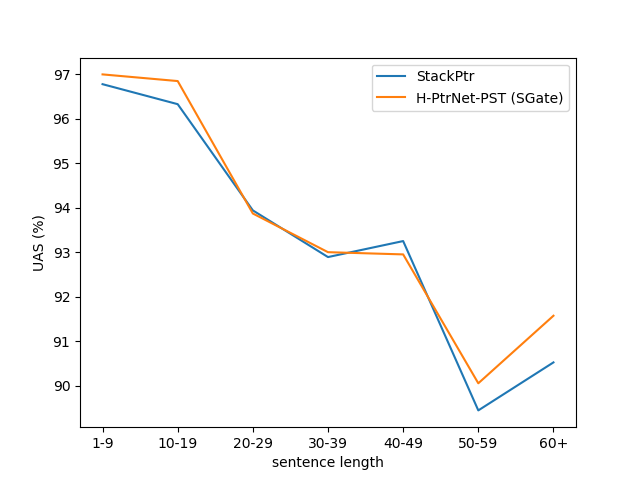}}
\caption{Performance analysis on French and Italian based on H-PtrNet-PST (SGate).}
\label{fig:dep_performance_more_examples}

\end{figure}


Figure \ref{fig:dependency_performance_2} shows the performance of our model with {\sc SGate} (Eq. \ref{eq:hptr_gate2}), where we can see that the performance on long sentences has been improved significantly. In the meanwhile, it still maintains higher accuracy than StackPtr on the short sentences (10 to 29 words). Figure \ref{fig:dep_performance_more_examples} shows two more examples, again, from which we can see that our model with {\sc SGate} tends to outperform StackPtr on longer sentences.



\subsection{Discourse Parsing}


\paragraph{Dataset.}
We use the standard RST Discourse Treebank (RST-DT) \cite{Carlson02}, which contains discourse annotations for 385 news articles from Penn Treebank \cite{Marcus93}. We evaluate our model in sentence-level parsing, for which we extract all the well-formed sentence-level discourse trees from document-level trees. In all, the training data contains 7321 sentences, and the testing data contains 951 sentences. These numbers match the statistics reported by \citet{Xiang19}. We follow the same settings as in their experiments and randomly choose 10\% of the training data for hyperparameter tuning.

\paragraph{Metric and Relation Labels.}
Following the standard in RST parsing, we use the unlabeled (Span) and labeled (Nuclearity, Relation) metrics proposed by \citet{Marcu00}. We only present $F_1$-score for space limitations. Following the previous work, we attach the nuclearity labels (NS, SN, NN) to 18 discourse relations, together giving 39 distinctive relation labels. 


\paragraph{Experimental Settings.} Since our goal is to evaluate our parsing method, we conduct the experiments based on gold EDU segmentations.  We compare our results with the recently proposed pointer network based parser of \citet{Xiang19} (Pointer Net). However, unlike their paper, we report results for both cases: ($i$) when the model was selected based on the best performance on Span identification; and ($ii$) when it was selected based on the relation labeling performance on the development set. We retrain their model for both settings. We also apply Adam optimizer as optimization algorithm and ELMo \cite{Peters:2018} with 0.5 dropout rate as word embeddings.    





\renewcommand{\arraystretch}{1.3}
\rowcolors{7}{white}{gray!15}
\begin{table}[t!]
\scalebox{0.68}{
\begin{tabular}{l|ccc}  
\textbf{Approach} & \multicolumn{1}{c}{\bf{Span}} & \bf{Nuclearity} & \bf{Relation}\\
\midrule
\bf{Human Agreement} & 95.7 & 90.4 & 83.0 \\ 
\midrule
\rowcolor{black!60}
\head{\bf{Baselines} }& & & \\

 \citet{Joty-2012} & 94.6  & 86.9  & 77.1  \\ 
 \citet{ji-eisenstein:2014:P14-1} & 93.5  & 81.3  & 70.5 \\ 
 \citet{Wang-acl-2017} & 95.6 & 87.8 & 77.6\\
\rowcolor{gray!15}
Pointer Net$^{\dag}$ \cite{Xiang19} & 97.39$_{\pm0.1}$ & 91.01$_{\pm0.4}$ & 81.08$_{\pm0.4}$ \\
\rowcolor{white}
Pointer Net$^{\S}$ \cite{Xiang19} & 97.14$_{\pm0.1}$ & 91.00$_{\pm0.3}$ & 81.29$_{\pm0.2}$ \\


\midrule
\rowcolor{black!60}
\head{\bf{Proposed Model} }& & & \\
H-PtrNet-P${^\dag}$ & \textbf{97.68$_{\pm0.03}$} & 91.86$_{\pm0.2}$ & 81.82$_{\pm0.2}$ \\

H-PtrNet-P${^\S}$ & 97.51$_{\pm0.08}$ & 91.98$_{\pm0.1}$ & 82.11$_{\pm0.2}$ \\

H-PtrNet-PS${^\dag}$ & 97.56$_{\pm0.1}$ & 91.52$_{\pm0.3}$ & 82.05$_{\pm0.5}$  \\
H-PtrNet-PS${^\S}$ & 97.35$_{\pm0.1}$ & 91.78$_{\pm0.1}$ & 82.35$_{\pm0.2}$  \\


H-PtrNet-PST${^\dag}$  & 97.56$_{\pm0.06}$ & 91.97$_{\pm0.3}$ & 82.37$_{\pm0.4}$ \\
H-PtrNet-PST${^\S}$  & 97.48$_{\pm0.2}$ & \textbf{92.01$_{\pm0.2}$} & \textbf{82.77$_{\pm0.2}$} \\
\bottomrule
\end{tabular}
}
\caption{Discourse parsing results with gold segmentation. ${^\dag}$ denotes the models selected based on Span. ${^\S}$ denotes the models selected based on Relation. We run all the experiments for three times and report the averages and the standard deviations.}
\label{table:pargold-results}
\end{table}  

\paragraph{Results.}
We present the results in Table \ref{table:pargold-results}. 
In discourse parsing, the number of EDUs in a sentence is relatively small compared to the sentence lengths (in words) in dependency parsing. Based on the observation in dependency parsing that the performance of H-PtrNet may drop for longer sentences due to parent error accumulation, we expect that in discourse parsing, this should not be the case since the the number of parsing steps is much smaller compared to that of dependency parsing.


We first consider the models that were selected based on Span performance (models with ${\dag}$ superscript). H-PtrNet-P, with only parent connection, outperforms the baseline in all three tasks. It achieves an absolute improvement of 0.29 $F_1$ in span identification compared to the baseline. Considering the performance has already exceeded the human agreement of 95.7 $F_1$, this gain is remarkable. Thanks to the higher accuracy on finding the right spans, we also achieve 0.85 and 0.74 absolute improvements in Nuclearity and Relation tasks, respectively. 
By adding the sibing and temporal connections, we test the performance of our full model, H-PtrNet-PST. The performance on Span is 0.17 $F_1$ higher than the baseline. However, it is not on par with our H-PtrNet-P. But, it is not surprising since we adopt binary tree structures in discourse parsing, which means the sibling information could be redundant in most cases. This also accords with our previous assumption that parent connections may bring enough information to decode RST trees.

Now we consider the models that were selected based on Relation labeling performance (models with ${\S}$ superscript). We achieve significant improvement in Relation compared to the baseline. Eventually the parser yields an $F_1$ of 82.77, which is very close to the human agreement (83.0 $F_1$). We observe that the performance in H-PtrNet-PS and H-PtrNet-PST is better than the H-PtrNet-P. {As the relation classifier and the pointer network share the same encoder information  (Sec. \ref{sec:model}), we believe that richer decoder information leads the model to learn better representations of the text spans (encoder states) and further leads to a better performance in relation labeling. } 


\begin{figure}[t]
\includegraphics[scale=0.24]{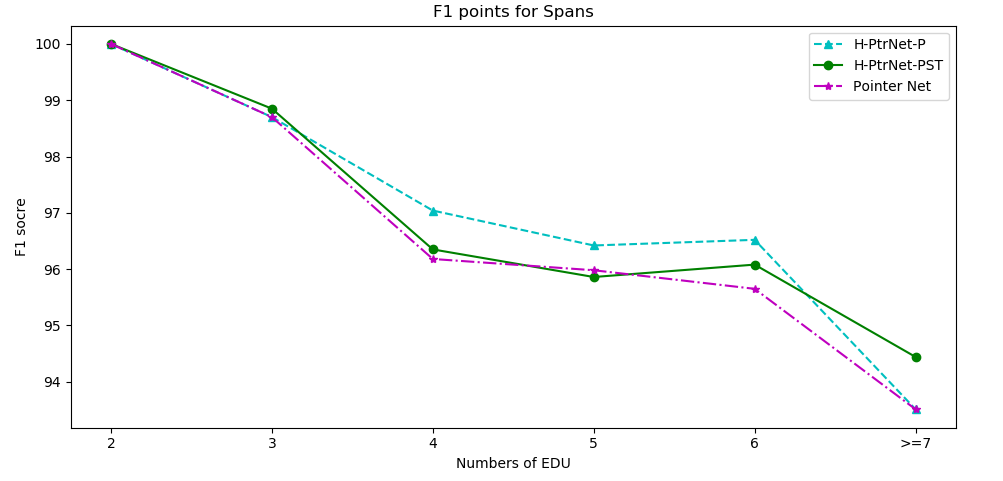} 
\caption{$F_1$ score in Span for different EDU numbers.}
\label{fig:discourseanalysis}
\end{figure}


We further analyze the performance of our proposed model in terms of number of EDUs. We present the $F_1$ score in Span of H-PtrNet-P and H-PtrNet-PST as well as the baseline in Figure \ref{fig:discourseanalysis}. It can be observed that both H-PtrNet-P and H-PtrNet-PST outperform the baseline with respect to almost every number of EDUs. Moreover, we can see that the H-PtrNet-P performs better in most of the cases, which once again conforms to our assumption that parent information is enough to decode RST trees. However, as discussed in our dependency parsing experiments, when the number of words (EDUs for discourse parsing) increases, the model may suffer from error accumulation from early parsing. Hence, H-PtrNet-PST tends to perform better when EDU number becomes large. 


\section{Conclusions} \label{sec:con}
In this paper, we propose hierarchical pointer network parsers and apply them to dependency and discourse  parsing tasks. Our parsers address the limitation of previous methods, where the decoder has a sequential structure while it is decoding a hierarchical tree structure, by allowing more flexible information flow to help the decoder receive the most relevant information.
For both tasks, our parsers outperform existing methods and set new state-of-the-arts of the two tasks. The broken-down analysis clearly illustrates that our parsers perform better for long sequences, complying with the motivation of our model. 

\section*{Acknowledgements}
This research is partly supported by the Alibaba-NTU Singapore Joint Research Institute, Nanyang Technological University (NTU), Singapore. Shafiq Joty would like to thank the funding support from his Start-up Grant (M4082038.020).  

\bibliography{parsing.bib}
\bibliographystyle{acl_natbib}

\end{document}


\appendix

\section{Appendix}
\subsection{Hyper-Parameters for Dependency Parsing}
We use the same Hyper-Parameters as StackPtr over all languages for dependency parsing. Table \ref{table:dep_hyp_param} summarizes all hyper parameters we are using.

\begin{table}[H]
\centering
\begin{tabular}{|l|l|l|}
\hline
\bf{Layer}                & \bf{Hyper-parameter}  & \bf{Value}      \\ \hline
\multirow{2}{*}{CNN}      & window size           & 3          \\ \cline{2-3} 
                          & number of filters     & 50         \\ \hline
\multirow{4}{*}{LSTM}     & encoder layers        & 3          \\ \cline{2-3} 
                          & encoder size          & 512        \\ \cline{2-3} 
                          & decoder layers        & 1          \\ \cline{2-3} 
                          & decoder size          & 512        \\ \hline
\multirow{2}{*}{MLP}      & arc MLP size          & 512        \\ \cline{2-3} 
                          & label MLP size        & 128        \\ \hline
\multirow{3}{*}{Dropout}  & embeddings            & 0.33       \\ \cline{2-3} 
                          & LSTM hidden states    & 0.33       \\ \cline{2-3} 
                          & LSTM layers           & 0.33       \\ \hline
\multirow{5}{*}{Learning} & Optimizer             & Adam       \\ \cline{2-3} 
                          & initial learning rate & 0.01       \\ \cline{2-3} 
                          & $(\beta_1, \beta_2)$  & (0.9, 0.9) \\ \cline{2-3} 
                          & decay rate            & 0.75       \\ \cline{2-3} 
                          & gradient clipping     & 5.0        \\ \hline
\end{tabular}
\caption{Hyper-parameters for all experiments}
\label{table:dep_hyp_param}
\end{table}

\subsection{UD Treebanks}
Table \ref{table:ud_corpora} shows the UD Treebank corpora used for language.
\begin{table}[H]
\centering
\begin{tabular}{|l|l|}
\hline
\bf{Language}  & \bf{Corpora} \\ \hline
Bulgarian & BTB     \\ \hline
Catalan   & AnCora  \\ \hline
English   & EWT     \\ \hline
French    & GSD     \\ \hline
German    & GSD     \\ \hline
Italian   & ISDT    \\ \hline
Romanian  & RRT     \\ \hline
\end{tabular}
\caption{UD Treebanks corpora of the 7 languages that we have tested.}
\label{table:ud_corpora}
\end{table}

\subsection{Hyper-Parameters for Discourse Parsing}
The hyper-parameters used for discourse parsing experiments are listed in table \ref{table:parsing-hyperparameters}.

\begin{table}[h]
\begin{center}
\begin{tabular}{|l|c|}
\hline \bf Hyper-parameters & \bf \ \ \ \ Value \ \ \ \\  \hline
Minibatch size & 64   \\
Embedding size & 1024 \\
Encoder hidden size & 64 \\
Decoder hidden size & 64 \\
ELMo dropout rate & 0.5 \\
Encoder dropout rate & 0.4 \\
Decoder dropout rate & 0.6 \\
Classifier dropout rate & 0.5 \\
Initial Learning rate & 0.001 \\
Adam $\beta_1$ & 0.9 \\
Adam $\beta_2$ & 0.95 \\
$L_2$ Regularization strength & 0.0005 \\

\hline
\end{tabular}
\end{center}
\caption{\label{table:parsing-hyperparameters} Optimal hyper-parameter settings for parser}
\end{table}
